\documentclass{article} 
\usepackage{iclr2024_conference,times}

\usepackage{amsmath,amsfonts,bm}

\def\eqref#1{equation~\ref{#1}}

\def\1{\bm{1}}

\DeclareMathAlphabet{\mathsfit}{\encodingdefault}{\sfdefault}{m}{sl}
\SetMathAlphabet{\mathsfit}{bold}{\encodingdefault}{\sfdefault}{bx}{n}

\usepackage{graphicx}
\usepackage{animate}

\usepackage{url}
\usepackage{xspace}
\usepackage{booktabs}
\usepackage{wrapfig}
\usepackage{multirow, multicol}
\usepackage{tabularx}
\usepackage{caption}
\usepackage{xcolor}
\usepackage{adjustbox}
\usepackage[accsupp]{axessibility}  
\definecolor{citeblue}{RGB}{48,111,186}
\usepackage[pagebackref=false,breaklinks=true,colorlinks=true,citecolor=citeblue,bookmarks=false]{hyperref}
\usepackage[capitalize]{cleveref}
\usepackage{tablefootnote}
\usepackage{float}

\title{AnimateDiff: Animate Your Personalized \\
Text-to-Image Diffusion Models without \\
Specific Tuning
}

\author{Yuwei Guo$^{1}$\quad
\stepcounter{footnote}Ceyuan Yang$^{2}$\thanks{Corresponding Author.}\quad
Anyi Rao$^{3}$\quad
Zhengyang Liang$^{2}$\quad
Yaohui Wang$^{2}$\\
\textbf{Yu Qiao$^{2}$\quad
Maneesh Agrawala$^{3}$\quad
Dahua Lin$^{1,2}$\quad
Bo Dai$^{2}$} \\
$^1$The Chinese University of Hong Kong\quad
$^2$Shanghai Artificial Intelligence Laboratory \\
$^3$Stanford University
}

\newcommand{\method}{AnimateDiff\xspace}
\newcommand{\methodlora}{MotionLoRA\xspace}

\newcommand{\numColumns}{3}
\newcommand{\columnSpacing}{0.25em}

\newcommand{\animationNotes}{
\textit{Best viewed with Acrobat Reader. Click the images to play the animation clips.}
}

\iclrfinalcopy 
\begin{document}

\maketitle

\begin{figure}[h]
    \begin{tabular}{
        @{}
        p{\dimexpr(\textwidth-\columnSpacing*(\numColumns-1))/\numColumns} @{\hspace{\columnSpacing}}
        p{\dimexpr(\textwidth-\columnSpacing*(\numColumns-1))/\numColumns} @{\hspace{\columnSpacing}}
        p{\dimexpr(\textwidth-\columnSpacing*(\numColumns-1))/\numColumns} @{}
    }
        \animategraphics[width=\linewidth]{8}{videos/teaser/qualitative_01/}{01}{16} & 
        \animategraphics[width=\linewidth]{8}{videos/teaser/qualitative_05/}{01}{16} & 
        \animategraphics[width=\linewidth]{8}{videos/teaser/qualitative_04/}{01}{16}
    \end{tabular}
    \begin{tabularx}{\textwidth}{XXX}
        \emph{\small \textbf{(cartoon)} 1boy, dark skin, playing guitar, concert, \ldots} & 
        \emph{\small \textbf{(oil painting)} black pearl pirate ship, night time, sea, \ldots} &
        \emph{\small \textbf{(realistic)} a Lamborghini on road, fireworks, high detail, \ldots}
    \end{tabularx}
    
    \renewcommand{\numColumns}{4}
    \renewcommand{\columnSpacing}{0.25em}
    \begin{tabular}{
        @{}
        p{\dimexpr(\textwidth-\columnSpacing*(\numColumns-1))/\numColumns} @{\hspace{\columnSpacing}}
        p{\dimexpr(\textwidth-\columnSpacing*(\numColumns-1))/\numColumns} @{\hspace{\columnSpacing}}
        p{\dimexpr(\textwidth-\columnSpacing*(\numColumns-1))/\numColumns} @{\hspace{\columnSpacing}}
        p{\dimexpr(\textwidth-\columnSpacing*(\numColumns-1))/\numColumns} @{}
    }
        \animategraphics[width=\linewidth]{8}{videos/teaser/lora_01/}{01}{16} & 
        \animategraphics[width=\linewidth]{8}{videos/teaser/lora_02/}{01}{16} & 
        \animategraphics[width=\linewidth]{8}{videos/teaser/lora_03/}{01}{16} & 
        \animategraphics[width=\linewidth]{8}{videos/teaser/lora_04/}{01}{16}
    \end{tabular}
    \begin{tabularx}{\textwidth}{XXXX}
        \centering \emph{\small \textbf{zoom-in}}            & 
        \centering \emph{\small \textbf{rolling}}            & 
        \centering \emph{\small \textbf{zoom-out + rolling}} &
        \centering \emph{\small \textbf{right + up}}
    \end{tabularx}
    
    \caption{
        \method directly turns existing personalized text-to-image (T2I) models to the corresponding animation generators with a pre-trained motion module.
        \emph{\textbf{First row}}: results by combining \method with three personalized T2Is in different domains;
        \emph{\textbf{Second row}}: results of further combining \method with \methodlora(s) to achieve shot type controls.
        \animationNotes
    }
    \vspace{1em}
    \label{fig:teaser}
\end{figure}

\begin{abstract}
With the advance of text-to-image (T2I) diffusion models (\textit{e.g.}, Stable Diffusion) and corresponding personalization techniques such as DreamBooth and LoRA, everyone can manifest their imagination into high-quality images at an affordable cost.
However, adding motion dynamics to existing high-quality personalized T2Is and enabling them to generate animations remains an open challenge.
In this paper, we present \method, a practical framework for animating personalized T2I models without requiring model-specific tuning.
At the core of our framework is a plug-and-play motion module that can be trained once and seamlessly integrated into any personalized T2Is originating from the same base T2I.
Through our proposed training strategy, the motion module effectively learns transferable motion priors from real-world videos.
Once trained, the motion module can be inserted into a personalized T2I model to form a personalized animation generator.
We further propose \methodlora, a lightweight fine-tuning technique for \method that enables a pre-trained motion module to adapt to new motion patterns, such as different shot types, at a low training and data collection cost.
We evaluate \method and \methodlora on several public representative personalized T2I models collected from the community.
The results demonstrate that our approaches help these models generate temporally smooth animation clips while preserving the visual quality and motion diversity.
Codes and pre-trained weights are available at \url{https://github.com/guoyww/AnimateDiff}.

\end{abstract}

\section{Introduction}
Text-to-image (T2I) diffusion models~\citep{nichol2021glide, ramesh2022hierarchical, saharia2022photorealistic, rombach2022high} have greatly empowered artists and amateurs to create visual content using text prompts.
To further stimulate the creativity of existing T2I models, lightweight personalization methods, such as DreamBooth~\citep{ruiz2023dreambooth} and LoRA~\citep{hu2021lora} have been proposed.
These methods enable customized fine-tuning on small datasets using consumer-grade hardware such as a laptop with an RTX3080, thereby allowing users to adapt a base T2I model to new domains and improve visual quality at a relatively low cost.
Consequently, a large community of AI artists and amateurs has contributed numerous personalized models on model-sharing platforms such as~\cite{civitai} and~\cite{huggingface}.
While these personalized T2I models can generate remarkable visual quality, their outputs are limited to static images.
On the other hand, the ability to generate animations is more desirable in real-world production, such as in the movie and cartoon industries.
In this work, we aim to directly transform existing high-quality personalized T2I models into animation generators without requiring model-specific fine-tuning, which is often impractical in terms of computation and data collection costs for amateur users.

We present \method, an effective pipeline for addressing the problem of animating personalized T2Is while preserving their visual quality and domain knowledge.
The core of \method is an approach for training a plug-and-play motion module that learns reasonable motion priors from video datasets, such as WebVid-10M~\citep{bain2021frozen}.
At inference time, the trained motion module can be directly integrated into personalized T2Is and produce smooth and visually appealing animations without requiring specific tuning.
The training of the motion module in \method consists of three stages.
Firstly, we fine-tune a domain adapter on the base T2I to align with the visual distribution of the target video dataset.
This preliminary step guarantees the motion module concentrates on learning the motion priors rather than pixel-level details from the training videos.
Secondly, we inflate the base T2I together with the domain adapter and introduce a newly initialized motion module for motion modeling.
We then optimize this module on videos while keeping the domain adapter and base T2I weights fixed.
By doing so, the motion module learns generalized motion priors and can, via module insertion, enable other personalized T2Is to generate smooth and appealing animations aligned with their personalized domains.
The third stage of \method, also dubbed as \methodlora, aims to adapt the pre-trained motion module to specific motion patterns with a small number of reference videos and training iterations.
We achieve this by fine-tuning the motion module with the aid of Low-Rank Adaptation (LoRA)~\citep{hu2021lora}.
Remarkably, adapting to a new motion pattern can be achieved with as few as 50 reference videos.
Moreover, a \methodlora model requires only approximately 30M of additional storage space, further enhancing the efficiency of model sharing.
This efficiency is particularly valuable for users who are unable to bear the expensive costs of pre-training but desire to fine-tune the motion module for specific effects.

We evaluate the performance of \method and \methodlora on a diverse set of personalized T2I models collected from model-sharing platforms \citep{civitai, huggingface}.
These models encompass a wide spectrum of domains, ranging from 2D cartoons to realistic photographs, thereby forming a comprehensive benchmark for our evaluation.
The results of our experiments demonstrate promising outcomes.
In practice, we also found that a Transformer~\citep{vaswani2017attention} architecture along the temporal axis is adequate for capturing appropriate motion priors.
We also demonstrate that our motion module can be seamlessly integrated with existing content-controlling approaches~\citep{zhang2023adding, mou2023t2i} such as ControlNet without requiring additional training, enabling \method for controllable animation generation.

In summary, (1) we present \method, a practical pipeline that enables the animation generation ability of any personalized T2Is without specific fine-tuning;
(2) we verify that a Transformer architecture is adequate for modeling motion priors, which provides valuable insights for video generation;
(3) we propose \methodlora, a lightweight fine-tuning technique to adapt pre-trained motion modules to new motion patterns;
(4) we comprehensively evaluate our approach with representative community models and compare it with both academic baselines and commercial tools such as~\cite{gen2} and~\cite{pikalab}.
Furthermore, we showcase its compatibility with existing works for controllable generation.

\section{Related Work}
\textbf{Text-to-image diffusion models.}
Diffusion models~\citep{ho2020denoising, dhariwal2021diffusion, song2020denoising} for text-to-image (T2I) generation~\citep{gu2022vector, mokady2023null, podell2023sdxl, ding2021cogview, zhou2022towards, ramesh2021zero, li2022upainting} have gained significant attention in both academic and non-academic communities recently.
GLIDE~\citep{nichol2021glide} introduced text conditions and demonstrated that incorporating classifier guidance leads to more pleasing results.
DALL-E2~\citep{ramesh2022hierarchical} improves text-image alignment by leveraging the CLIP~\citep{radford2021learning} joint feature space.
Imagen~\citep{saharia2022photorealistic} incorporates a large language model~\citep{raffel2020exploring} and a cascade architecture to achieve photorealistic results.
Latent Diffusion Model~\citep{rombach2022high}, also known as Stable Diffusion, moves the diffusion process to the latent space of an auto-encoder to enhance efficiency.
eDiff-I~\citep{balaji2022ediffi} employs an ensemble of diffusion models specialized for different generation stages.

\textbf{Personalizing T2I models.}
To facilitate the creation with pre-trained T2Is, many works focus on efficient model personalization~\citep{shi2023instantbooth, lu2023specialist, dong2022dreamartist, kumari2023multi}, \emph{i.e.}, introducing concepts or styles to the base T2I using reference images.
The most straightforward approach to achieve this is complete fine-tuning of the model.
Despite its potential to significantly enhance overall quality, this practice can lead to catastrophic forgetting~\citep{kirkpatrick2017overcoming, french1999catastrophic} when the reference image set is small.
Instead, DreamBooth~\citep{ruiz2023dreambooth} fine-tunes the entire network with preservation loss and uses only a few images.
Textual Inversion~\citep{gal2022image} optimize a token embedding for each new concept.
Low-Rank Adaptation (LoRA)~\citep{hu2021lora} facilitates the above fine-tuning process by introducing additional LoRA layers to the base T2I and optimizing only the weight residuals.
There are also encoder-based approaches that address the personalization problem~\citep{gal2023designing, jia2023taming}.
In our work, we focus on tuning-based methods, including overall fine-tuning, DreamBooth~\citep{ruiz2023dreambooth}, and LoRA~\citep{hu2021lora}, as they preserve the original feature space of the base T2I.

\textbf{Animating personalized T2Is.}
There are not many existing works regarding animating personalized T2Is.
Text2Cinemagraph~\citep{mahapatra2023textguided} proposed to generate cinematography via flow prediction.
In the field of video generation, it is common to extend a pre-trained T2I with temporal structures.
Existing works~\citep{esser2023structure, zhou2022magicvideo, singer2022make, ho2022video, ho2022imagen, ruan2023mm, luo2023videofusion, yin2023nuwa, yin2023dragnuwa, wang2023lavie, hong2022cogvideo, luo2023videofusion} mostly update all parameters and modify the feature space of the original T2I and is not compatible with personalized ones.
Align-Your-Latents~\citep{blattmann2023align} shows that the frozen image layers in a general video generator can be personalized.
Recently, some video generation approaches have shown promising results in animating a personalized T2I model.
Tune-a-Video~\citep{wu2022tune} fine-tune a small number of parameters on a single video.
Text2Video-Zero~\citep{khachatryan2023text2video} introduces a training-free method to animate a pre-trained T2I via latent wrapping based on a pre-defined affine matrix.

\section{Preliminary}\label{sec:preliminary}

We introduce the preliminary of Stable Diffusion~\citep{rombach2022high}, the base T2I model used in our work, and Low-Rank Adaptation (LoRA)~\citep{hu2021lora}, which helps understand the domain adapter (Sec.~\ref{method:adapter}) and \methodlora (Sec.~\ref{method:motion_lora}) in \method.

\textbf{Stable Diffusion.}
We chose Stable Diffusion (SD) as the base T2I model in this paper since it is open-sourced and has a well-developed community with many high-quality personalized T2I models for evaluation.
SD performs the diffusion process within the latent space of a pre-trained autoencoder $\mathcal{E}(\cdot)$ and $\mathcal{D}(\cdot)$.
In training, an encoded image $z_0 = \mathcal{E}(x_0)$ is perturbed to $z_t$ by the forword diffusion:
\begin{equation}\label{eq:noise}
    z_t = \sqrt{\bar{\alpha_t}}z_0 + \sqrt{1-\bar{\alpha_t}}\epsilon,~\epsilon\sim\mathcal{N}(0, \mathit{I}),
\end{equation}
for $t = 1,\ldots, T$, where pre-defined $\bar{\alpha_t}$ determines the noise strength at step $t$.
The denoising network $\epsilon_\theta(\cdot)$ learns to reverse this process by predicting the added noise, encouraged by an MSE loss:
\begin{equation}\label{eq:objective}
    \mathcal{L} = \mathbb{E}_{\mathcal{E}(x_0), y, \epsilon\sim\mathcal{N}(0, \mathit{I}), t}\left\lbrack 
\lVert \epsilon - \epsilon_\theta(z_t, t, \tau_\theta(y)) \rVert_2^2 \right\rbrack,
\end{equation}
where $y$ is the text prompt corresponding to $x_0$; $\tau_\theta(\cdot)$ is a text encoder mapping the prompt to a vector sequence.
In SD, $\epsilon_\theta(\cdot)$ is implemented as a UNet~\citep{ronneberger2015unet} consisting of pairs of down/up sample blocks at four resolution levels, as well as a middle block.
Each network block consists of ResNet~\citep{he2016deep}, spatial self-attention layers, and cross-attention layers that introduce text conditions.
\textbf{Low-rank adaptation (LoRA).}
LoRA~\citep{hu2021lora} is an approach that accelerates the fine-tuning of large models and is first proposed for language model adaptation.
Instead of retraining all model parameters, LoRA adds pairs of rank-decomposition matrices and optimizes only these newly introduced weights.
By limiting the trainable parameters and keeping the original weights frozen, LoRA is less likely to cause catastrophic forgetting~\citep{kirkpatrick2017overcoming}.
Concretely, the rank-decomposition matrices serve as the residual of the pre-trained model weights $\mathcal{W} \in \mathbb{R}^{m \times n}$.
The new model weight with LoRA is
\begin{equation}
    \mathcal{W}' = \mathcal{W} + \Delta \mathcal{W} = \mathcal{W} + AB^T,
\end{equation}
where $A \in \mathbb{R}^{m \times r}$, $B \in \mathbb{R}^{n \times r}$ are a pair of rank-decomposition matrices, $r$ is a hyper-parameter, which is referred to as the rank of LoRA layers.
In practice, LoRA is only applied to attention layers, further reducing the cost and storage for model fine-tuning.

\section{AnimateDiff}\label{sec:animatediff}

\begin{wrapfigure}{r}{6cm}
    \vspace{-1em}
    \includegraphics[width=\linewidth]{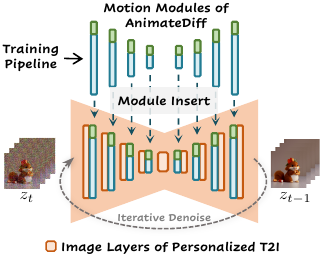}
    \vspace{-1.5em}
    \caption{Inference pipeline.}
    \vspace{-1em}
    \label{fig:overview}
\end{wrapfigure}

The core of our method is learning transferable motion priors from video data, which can be applied to personalized T2Is without specific tuning.
As shown in~\cref{fig:overview}, at inference time, our motion module (blue) and the \emph{optional} \methodlora (green) can be directly inserted into a personalized T2I to constitute the animation generator, which subsequently generates animations via an iterative denoising process.

We achieve this by training three components of \method, namely domain adapter, motion module, and \methodlora.
The domain adapter in Sec.~\ref{method:adapter} is only used in the training to alleviate the negative effects caused by the visual distribution gap between the base T2I pre-training data and our video training data;
the motion module in Sec.~\ref{method:module} is for learning the motion priors;
and the \methodlora in Sec.~\ref{method:motion_lora}, which is \emph{optional} in the case of general animation, is for adapting pre-trained motion modules to new motion patterns.
Sec.\ref{method:practice} elaborates on the training (\cref{fig:pipeline}) and inference of \method.

\subsection{Alleviate Negative Effects from Training Data with Domain Adapter}\label{method:adapter}
Due to the difficulty in collection, the visual quality of publicly available video training datasets is much lower than their image counterparts.
For example, the contents of the video dataset WebVid~\citep{bain2021frozen} are mostly real-world recordings, whereas the image dataset LAION-Aesthetic~\citep{schuhmann2022laion} contains higher-quality contents, including artistic paintings and professional photography.
Moreover, when treated individually as images, each video frame can contain motion blur, compression artifacts, and watermarks.
Therefore, there is a non-negligible quality domain gap between the high-quality image dataset used to train the base T2I and the target video dataset we use for learning the motion priors. 
We argue that such a gap can limit the quality of the animation generation pipeline when trained directly on the raw video data.

\begin{figure}
    \centering
    \includegraphics[width=\textwidth]{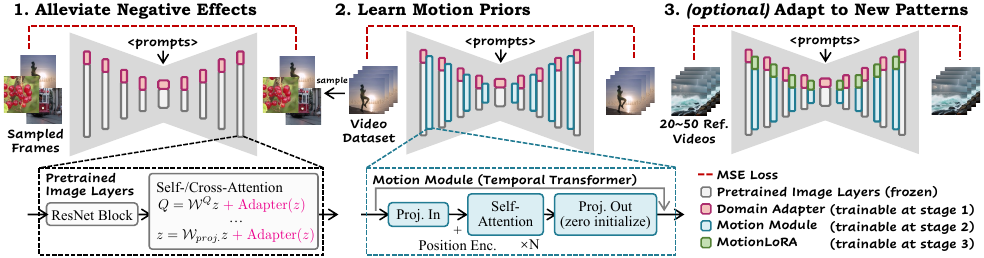}
    \caption{
        \textbf{Training pipeline of \method.}
        \method consists of three training stages for the corresponding component modules.
        Firstly, a domain adapter (Sec.~\ref{method:adapter}) is trained to alleviate the negative effects caused by training videos.
        Secondly, a motion module (Sec.~\ref{method:module}) is inserted and trained on videos to learn general motion priors.
        Lastly, \methodlora (Sec.~\ref{method:motion_lora}) is trained on a few reference videos to adapt the pre-trained motion module to new motion patterns.
    }
    \label{fig:pipeline}
\end{figure}

To avoid learning this quality discrepancy as part of our motion module and preserve the knowledge of the base T2I, 
we propose to \textbf{fit the domain information to a separate network}, dubbed as domain adapter.
We drop the domain adapter at inference time and show that this practice helps reduce the negative effects caused by the domain gap mentioned above.
We implement the domain adapter layers with LoRA~\citep{hu2021lora} and insert them into the self-/cross-attention layers in the base T2I, as shown in~\cref{fig:pipeline}.
Take query (Q) projection as an example. The internal feature $z$ after projection becomes
\begin{equation} \label{eq:domain_adapter}
    Q = \mathcal{W}^Qz + \text{AdapterLayer}(z) = \mathcal{W}^Qz + \alpha \cdot AB^Tz,
\end{equation}
where $\alpha = 1$ is a scalar and can be adjusted to other values at inference time (set to $0$ to remove the effects of domain adapter totally).
We then optimize only the parameters of the domain adapter on static frames randomly sampled from video datasets with the same objective in~\cref{eq:objective}.

\subsection{Learn Motion Priors with Motion Module}\label{method:module}

To model motion dynamics along the temporal dimension on top of a pre-trained T2I, we must 1) inflate the 2-dimensional diffusion model to deal with 3-dimensional video data and 2) design a sub-module to enable efficient information exchange along the temporal axis.
\textbf{Network Inflation.}
The pre-trained image layers in the base T2I model capture high-quality content priors.
To utilize the knowledge, a preferable way for network inflation is to let these image layers independently deal with video frames.
To achieve this, we adopt a practice similar to recent works~\citep{ho2022video, wu2022tune, blattmann2023align},
and modify the model so that it takes 5D video tensors $x \in \mathbb{R}^{b \times c \times f \times h \times w}$ as input, where $b$ and $f$ represent batch axis and frame-time axis respectively.
When the internal feature maps go through image layers, the temporal axis $f$ is ignored by being reshaped into the $b$ axis, allowing the network to process each frame independently.
We then reshape the feature map to the 5D tensor after the image layer.
On the other hand, our newly inserted motion module ignores the spatial axis by reshaping $h, w$ into $b$ and then reshaping back after the module.

\textbf{Module Design.}
Recent works on video generation have explored many designs for temporal modeling.
In \method, we adopt
the Transformer~\citep{vaswani2017attention} architecture as our motion module design, and make minor modifications to adapt it to operate along the temporal axis, which we refer to as ``temporal Transformer" in the following sections.
We experimentally found this design is adequate for modeling motion priors.
As illustrated in~\cref{fig:pipeline}, the temporal Transformer consists of several self-attention blocks along the temporal axis, with sinusoidal position encoding to encode the location of each frame in the animation.
As mentioned above, the input of the motion module is the reshaped feature map whose spatial dimensions are merged into the batch axis.
When we divide the reshaped feature map along the temporal axis, 
it can be regarded as vector sequences with length of $f$, \textit{i.e.}, $\{z_1, ..., z_f; z_i \in \mathbb{R}^{(b \times h \times w) \times c}\}$.
The vectors will then be projected and go through several self-attention blocks, \emph{i.e.}
\begin{equation}
    z_{out} = \text{Attention}(Q,K,V)=\text{Softmax}(QK^T / \sqrt{c})\cdot V,
\end{equation}
where $Q = W^Qz$, $K = W^Kz$, and $V=W^Vz$ are three separated projections.
The attention mechanism enables the generation of the current frame to incorporate information from other frames.
As a result, instead of generating each frame individually, the T2I model inflated with our motion module learns to capture the changes of visual content over time, which constitute the motion dynamics in an animation clip.
Note that sinusoidal position encoding added before the self-attention is essential; otherwise, the module is not aware of the frame order in the animation.
To avoid any harmful effects that the additional module might introduce, we zero initialize~\citep{zhang2023adding} the output projection layers of the temporal Transformer and add a residual connection so that the motion module is an identity mapping at the beginning of training.

\subsection{Adapt to New Motion Patterns with \methodlora}\label{method:motion_lora}
While the pre-trained motion module captures general motion priors, a question arises when we need to effectively adapt it to new motion patterns such as camera zooming, panning and rolling, \emph{etc.}, with a small number of reference videos and training iterations.
Such efficiency is essential for users who cannot afford expensive pre-training costs but would like to fine-tune the motion module for specific effects.
Here comes the last stage of \method, also dubbed as \methodlora (\cref{fig:pipeline}), an efficient fine-tuning approach for \textbf{motion personalization}.
Considering the architecture of the motion module and the limited number of reference videos, we add LoRA layers to the self-attention layers of the motion module in the inflated model described in Sec.~\ref{method:module}, then train these LoRA layers on the reference videos of new motion patterns.

We experiment with several shot types and get the reference videos via rule-based data augmentation.
For instance, to get videos with zooming effects, we augment the videos by gradually reducing (zoom-in) or enlarging (zoom-out) the cropping area of video frames along the temporal axis.
We demonstrate that our \methodlora can achieve promising results even with as few as $20\sim50$ reference videos, 2,000 training iterations (around $1\sim2$ hours) as well as about 30M storage space, enabling efficient model tuning and sharing among users.
Benefited by the low-rank property, \methodlora also has the composition capability.
Namely, individually trained \methodlora models can be combined to achieve composed motion effects at inference time.

\subsection{\method in Practice}\label{method:practice}
We elaborate on the training and inference here and put the detailed configurations in supplementary materials.

\textbf{Training.} 
As illustrated in~\cref{fig:pipeline}, \method consists of three trainable component modules to learn transferable motion priors.
Their training objectives are slightly different.
The domain adapter is trained with the original objective as in~\cref{eq:objective}.
The motion module and \methodlora, as part of an animation generator, use a similar objective with minor modifications to accommodate higher dimension video data.
Concretely, a video data batch $x_0^{1:f} \in \mathbb{R}^{b \times c \times f \times h \times w}$ is first encoded into the latent codes $z_0^{1:f}$ frame-wisely via the pre-trained auto-encoder of SD.
The latent codes are then noised using the defined forward diffusion schedule as in~\cref{eq:noise}
\begin{equation}
    z_t^{1:f} = \sqrt{\bar{\alpha_t}}z_0^{1:f} + \sqrt{1-\bar{\alpha_t}}\epsilon^{1:f}.
\end{equation}
The inflated model inputs the noised latent codes and corresponding text prompts and predicts the added noises.
The final training objective of our motion modeling module is:
\begin{equation}
    \mathcal{L} = \mathbb{E}_{\mathcal{E}(x_0^{1:f}), y, \epsilon^{1:f}\sim\mathcal{N}(0, \mathit{I}), t}\left\lbrack 
\lVert \epsilon - \epsilon_\theta(z_t^{1:f}, t, \tau_\theta(y)) \rVert_2^2 \right\rbrack.
\end{equation}

It's worth noting that when training the domain adapter, the motion module, and the \methodlora, parameters outside the trainable part remain frozen.

\textbf{Inference.}
At inference time (\cref{fig:overview}), the personalized T2I model will first be inflated in the same way discussed in~\cref{method:module}, then injected with the motion module for general animation generation, and the \emph{optional} \methodlora for generating animation with personalized motion.
As for the domain adapter, instead of simply dropping it during the inference time, in practice, we can also inject it into the personalized T2I model and adjust its contribution by changing the scaler $\alpha$ in \cref{eq:domain_adapter}.
An ablation study on the value of $\alpha$ is conducted in experiments.
Finally, the animation frames can be obtained by performing the reverse diffusion process and decoding the latent codes.

\section{Experiments}

We implement \method upon Stable Diffusion V1.5 and train motion module using the WebVid-10M~\citep{bain2021frozen} dataset.
Detailed configurations can be found in supplementary materials.

\subsection{Qualitative Results}

\textbf{Evaluate on community models.}
We evaluated the \method with a diverse set of representative personalized T2Is collected from~\cite{civitai}.
These personalized T2Is encompass a wide range of domains, thus serving as a comprehensive benchmark.
Since personalized domains in these T2Is only respond to certain ``trigger words", we abstain from using common text prompts but refer to the model homepage to construct the evaluation prompts.
In~\cref{fig:qualitative}, we show eight qualitative results of \method.
Each sample corresponds to a distinct personalized T2I.
In the second row of Figure \ref{fig:teaser}, we present the outcomes obtained by integrating \method with \methodlora to achieve shot type controls.
The last two samples exhibit the composition capability of \methodlora, achieved by linearly combining the individually trained weights.

\begin{figure}[t]
    \begin{tabularx}{\textwidth}{XXXX}
        \centering \emph{\small \textbf{RCNZ Cartoon 3d}} & 
        \centering \emph{\small \textbf{TUSUN}} & 
        \centering \emph{\small \textbf{epiC Realism}} & 
        \centering \emph{\small \textbf{ToonYou}}
    \end{tabularx}
    \renewcommand{\numColumns}{4}
    \renewcommand{\columnSpacing}{0.25em}
    \begin{tabular}{
        @{}
        p{\dimexpr(\textwidth-\columnSpacing*(\numColumns-1))/\numColumns} @{\hspace{\columnSpacing}}
        p{\dimexpr(\textwidth-\columnSpacing*(\numColumns-1))/\numColumns} @{\hspace{\columnSpacing}}
        p{\dimexpr(\textwidth-\columnSpacing*(\numColumns-1))/\numColumns} @{\hspace{\columnSpacing}}
        p{\dimexpr(\textwidth-\columnSpacing*(\numColumns-1))/\numColumns} @{}
    }
        \animategraphics[width=\linewidth]{8}{videos/qualitative/05/}{01}{16} & 
        \animategraphics[width=\linewidth]{8}{videos/qualitative/06/}{01}{16} & 
        \animategraphics[width=\linewidth]{8}{videos/qualitative/03/}{01}{16} & 
        \animategraphics[width=\linewidth]{8}{videos/qualitative/01/}{01}{16}
    \end{tabular}
    \vspace{0.5em}
    \begin{tabularx}{\textwidth}{XXXX}
        \emph{\small a golden Labrador, natural lighting, \ldots} & 
        \emph{\small cute Pallas's Cat walking in the snow, \ldots} & 
        \emph{\small photo of 24 y.o woman, night street, \ldots} &
        \emph{\small coastline, lighthouse, waves, sunlight, \ldots} 
    \end{tabularx}
    \begin{tabularx}{\textwidth}{XXXX}
        \centering \emph{\small \textbf{MeinaMix}} & 
        \centering \emph{\small \textbf{Realistic Vision}} & 
        \centering \emph{\small \textbf{MoXin}} & 
        \centering \emph{\small \textbf{Oil painting}}
    \end{tabularx}
    \renewcommand{\numColumns}{4}
    \renewcommand{\columnSpacing}{0.25em}
    \begin{tabular}{
        @{}
        p{\dimexpr(\textwidth-\columnSpacing*(\numColumns-1))/\numColumns} @{\hspace{\columnSpacing}}
        p{\dimexpr(\textwidth-\columnSpacing*(\numColumns-1))/\numColumns} @{\hspace{\columnSpacing}}
        p{\dimexpr(\textwidth-\columnSpacing*(\numColumns-1))/\numColumns} @{\hspace{\columnSpacing}}
        p{\dimexpr(\textwidth-\columnSpacing*(\numColumns-1))/\numColumns} @{}
    }
        \animategraphics[width=\linewidth]{8}{videos/qualitative/07/}{01}{16} & 
        \animategraphics[width=\linewidth]{8}{videos/qualitative/04/}{01}{16} & 
        \animategraphics[width=\linewidth]{8}{videos/qualitative/08/}{01}{16} & 
        \animategraphics[width=\linewidth]{8}{videos/qualitative/02/}{01}{16}
    \end{tabular}
    \begin{tabularx}{\textwidth}{XXXX}
        \emph{\small 1girl, white hair, purple eyes, dress, petals, \ldots} &
        \emph{\small a cyberpunk city street, night time, \ldots} &
        \emph{\small a bird sits on a branch, ink painting, \ldots} &
        \emph{\small sunset, orange sky, fishing boats, waves, \ldots} 
    \end{tabularx}
    \caption{
        \textbf{Qualitative Result.}
        Each sample corresponds to a distinct personalized T2I.
        \animationNotes
    }
    \label{fig:qualitative}
\end{figure}

\textbf{Compare with baselines.}
In the absence of existing methods specifically designed for animating personalized T2Is, we compare our method with two recent works in video generation that can be adapted for this task: \textbf{1) Text2Video-Zero}~\citep{khachatryan2023text2video} and \textbf{2) Tune-a-Video} \citep{wu2022tune}.
We also compare \method with two commercial tools: \textbf{3)~\cite{gen2}} for text-to-video generation, and \textbf{4)~\cite{pikalab}} for image animation.
The results are shown in~\cref{fig:qualitative_compare}.

\begin{figure}[t]
    \begin{tabularx}{\textwidth}{XXXX}
        \centering Tune-A-Video &
        \centering \method & 
        \centering T2V-Zero &
        \centering \method
    \end{tabularx}
    \vspace{-0.25em}
    \renewcommand{\numColumns}{4}
    \renewcommand{\columnSpacing}{0.25em}
    \begin{tabular}{
        @{}
        p{\dimexpr(\textwidth-\columnSpacing*(\numColumns-1))/\numColumns} @{\hspace{\columnSpacing}}
        p{\dimexpr(\textwidth-\columnSpacing*(\numColumns-1))/\numColumns} @{\hspace{\columnSpacing}}
        p{\dimexpr(\textwidth-\columnSpacing*(\numColumns-1))/\numColumns} @{\hspace{\columnSpacing}}
        p{\dimexpr(\textwidth-\columnSpacing*(\numColumns-1))/\numColumns} @{}
    }
        \animategraphics[width=\linewidth]{8}{videos/qualitative_compare/baseline_01/01/baseline/}{01}{16} & 
        \animategraphics[width=\linewidth]{8}{videos/qualitative_compare/baseline_01/01/ours/}{01}{16}     & 
        \animategraphics[width=\linewidth]{8}{videos/qualitative_compare/baseline_02/01/baseline/}{01}{16} & 
        \animategraphics[width=\linewidth]{8}{videos/qualitative_compare/baseline_02/01/ours/}{01}{16}
    \end{tabular}
    \begin{tabularx}{\textwidth}{XX}
        \centering \emph{\small a raccoon is playing guitar, soft lighting, \ldots} & 
        \centering \emph{\small a horse galloping on the street, best quality, \ldots}
    \end{tabularx}

    \caption{
        \textbf{Qualitative Comparison.}
        \animationNotes
    }
    \label{fig:qualitative_compare}
\end{figure}

\subsection{Quantitative Comparison}

We conduct the quantitative comparison through user study and CLIP metrics.
The comparison focuses on three key aspects: \emph{text alignment}, \emph{domain similarity}, and \emph{motion smoothness}.
The results are shown in~\cref{tab:quantitative}.
Detailed implementations can be found in supplementary materials.

\textbf{User study.}
In the user study, we generate animations using all three methods based on the same personalized T2I models.
Participants are then asked to individually rank the results based on the above three aspects.
We use the Average User Ranking (AUR) as a preference metric where a higher score indicates superior performance.
Note that the corresponding prompts and images are provided for reference for text alignment and domain similarity evaluation.

\textbf{CLIP metric.}
We also employed the CLIP~\citep{radford2021learning} metric, following the approach taken by previous studies~\citep{wu2022tune, khachatryan2023text2video}.
When evaluating domain similarity, it is important to note that the CLIP score was computed between the animation frames and the reference images generated using the personalized T2Is.

\begin{table}
    \centering
    \caption{Quantitative comparison.
    A higher score indicates superior performance.}
    \vspace{-0.8em}
    \begin{tabular}{lcccccccc}
        \bottomrule
        \multirow{2}{*}{Method} & & \multicolumn{3}{c}{\textbf{User Study ($\uparrow$)}} & & \multicolumn{3}{c}{{\textbf{CLIP Metric ($\uparrow$)}}} \\
            & & Text. & Domain. & Smooth. & & Text. & Domain. & Smooth. \\
        \hline
            Text2Video-Zero & & 1.620          & \textbf{2.620} & 1.560          & & 32.04          & 84.84          & 96.57 \\
            Tune-a-Video    & & 2.180          & 1.100          & 1.615          & & \textbf{35.98} & 80.68          & 97.42 \\
            \textbf{Ours}   & & \textbf{2.210} & 2.280          & \textbf{2.825} & & 31.39          & \textbf{87.29} & \textbf{98.00} \\
        \toprule
    \end{tabular}
    \label{tab:quantitative}
    \vspace{-1em}
\end{table}

\subsection{Ablative Study}

\begin{figure}[t]
    \centering
    \includegraphics[width=\linewidth]{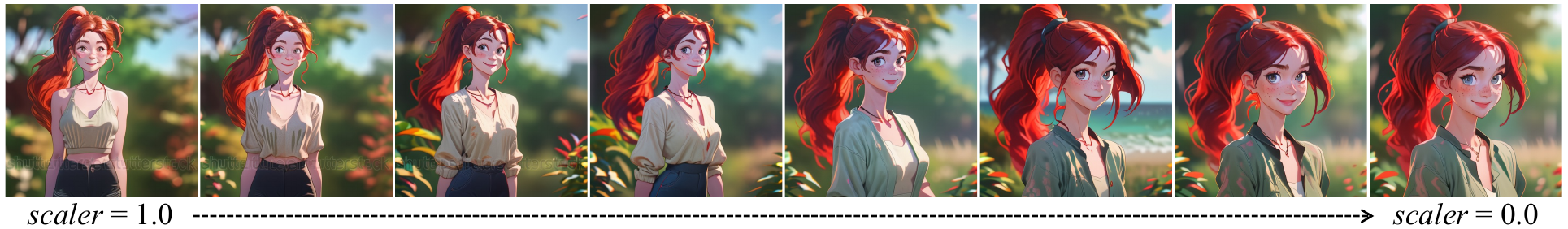}
    \caption{
    \textbf{Ablation on domain adapter.}
    We adjust the scaler of the adapter from $1$ to $0$ to gradually remove its effects.
    In this figure, we show the first frame of the generated animation.
    }
    \label{fig:ablation_eraser}
\end{figure}

\textbf{Domain adapter.}
To investigate the impact of the domain adapter in \method, we conducted a study by adjusting the scaler in the adapter layers during inference, ranging from $1$ (full impact) to $0$ (complete removal).
As illustrated in Figure~\ref{fig:ablation_eraser}, as the scaler of the adapter decreases, there is an improvement in overall visual quality, accompanied by a reduction in the visual content distribution learned from the video dataset (the watermark in the case of WebVid~\citep{bain2021frozen}).
These results indicate the successful role of the domain adapter in enhancing the visual quality of \method by alleviating the motion module from learning the visual distribution gap.

\textbf{Motion module design.}
We compare our motion module design of the temporal Transformer with its full convolution counterpart, which is motivated by the fact that both designs are widely employed in recent works on video generation.
We replace the temporal attention with 1D temporal convolution and ensured that the two model parameters were closely aligned.
As depicted in supplementary materials, the convolutional motion module aligns all frames to be identical but does not incorporate any motion compared to the Transformer architecture.

\begin{figure}
    \renewcommand{\numColumns}{5}
    \renewcommand{\columnSpacing}{0.1em}
    \begin{tabular}{
        @{}
        p{\dimexpr(\linewidth-\columnSpacing*(\numColumns+2))/\numColumns} @{\hspace{\columnSpacing}}
        p{\dimexpr(\linewidth-\columnSpacing*(\numColumns+2))/\numColumns} @{\hspace{\dimexpr\columnSpacing*4}} 
        p{\dimexpr(\linewidth-\columnSpacing*(\numColumns+2))/\numColumns} @{\hspace{\columnSpacing}}
        p{\dimexpr(\linewidth-\columnSpacing*(\numColumns+2))/\numColumns} @{\hspace{\columnSpacing}}
        p{\dimexpr(\linewidth-\columnSpacing*(\numColumns+2))/\numColumns} @{}
    }
        \animategraphics[width=\linewidth]{8}{videos/ablation_study/rank/02/}{01}{16} & 
        \animategraphics[width=\linewidth]{8}{videos/ablation_study/rank/01/}{01}{16} & 
        \animategraphics[width=\linewidth]{8}{videos/ablation_study/num_video/01/}{01}{16} & 
        \animategraphics[width=\linewidth]{8}{videos/ablation_study/num_video/02/}{01}{16} & 
        \animategraphics[width=\linewidth]{8}{videos/ablation_study/num_video/03/}{01}{16}
    \end{tabular}
    \begin{tabularx}{\linewidth}{@{}X@{\hspace{\columnSpacing}}X@{\hspace{\columnSpacing*4}}X@{\hspace{\columnSpacing}}X@{\hspace{\columnSpacing}}X@{}
    }
       \centering \emph{\small rank=2 ($\sim$1M)} & 
       \centering \emph{\small rank=128 ($\sim$36M)} & 
       \centering \emph{\small N=5} & 
       \centering \emph{\small N=50} & 
       \centering \emph{\small N=1000}
    \end{tabularx}
    \caption{
    \textbf{Ablation on MotionLoRA's efficiency.}
    \emph{Two samples on the left}: with different network rank;
    \emph{Three samples on the right}: with different numbers of reference videos.
    \animationNotes
    }
    \label{fig:efficiency_motionlora}
\end{figure}

\textbf{Efficiency of \methodlora.}
The efficiency of \methodlora in \method was examined in terms of \textit{parameter efficiency} and \textit{data efficiency}.
Parameter efficiency is crucial for efficient model training and sharing among users, while data efficiency is essential for real-world applications where collecting an adequate number of reference videos for specific motion patterns may be challenging.
To investigate these aspects, we trained multiple \methodlora models with varying parameter scales and reference video quantities.
In~\cref{fig:efficiency_motionlora}, the first two samples demonstrate that \methodlora is capable of learning new camera motions (\emph{e.g.}, zoom-in) with a small parameter scale while maintaining comparable motion quality.
Furthermore, even with a modest number of reference videos (\emph{e.g.}, $N=50$), the model successfully learns the desired motion patterns.
However, when the number of reference videos is excessively limited (\emph{e.g.}, $N=5$), significant degradation in quality is observed, suggesting that \methodlora encounters difficulties in learning shared motion patterns and instead relies on capturing texture information from the reference videos.

\subsection{Controllable generation.}

\begin{wrapfigure}{r}{0.5\textwidth}
    \vspace{-1em}
    \animategraphics[width=\linewidth]{8}{videos/control/01/}{01}{16}
    \emph{\small city street, neon, fog, closeup portrait photo of young woman in dark clothes, \ldots}
    \caption{
    \textbf{Controllable generation.}
    \animationNotes
    }
    \vspace{-1em}
    \label{fig:control}
\end{wrapfigure}

The separated learning of visual content and motion priors in \method enables the direct application of existing content control approaches for controllable generation.
To demonstrate this capability, we combined \method with ControlNet~\citep{zhang2023adding} to control the generation with extracted depth map sequence.
In contrast to recent video editing techniques~\citep{ceylan2023pix2video, wang2023zero} that employ DDIM~\citep{song2020denoising} inversion to obtain smoothed latent sequences, we generate animations from randomly sampled noise.
As illustrated in Figure \ref{fig:control}, our results exhibit meticulous motion details (such as hair and facial expressions) and high visual quality.

\section{Conclusion}
In this paper, we present \method, a practical pipeline directly turning personalized text-to-image (T2I) models for animation generation once and for all, without compromising quality or losing pre-learned domain knowledge.
To accomplish this, we design three component modules in \method to learn meaningful motion priors while alleviating visual quality degradation and enabling motion personalization with a lightweight fine-tuning technique named \methodlora.
Once trained, our motion module can be integrated into other personalized T2Is to generate animated images with natural and coherent motions while remaining faithful to the personalized domain.
Extensive evaluation with various personalized T2I models also validates the effectiveness and generalizability of our \method and \methodlora.
Furthermore, we demonstrate the compatibility of our method with existing content-controlling approaches, enabling controllable generation without incurring additional training costs.
Overall, \method provides an effective baseline for personalized animation and holds significant potential for a wide range of applications.

\section{Ethics Statement}
We strongly condemn the misuse of generative AI to create content that harms individuals or spreads misinformation.
However, we acknowledge the potential for our method to be misused since it primarily focuses on animation and can generate human-related content.
It is also important to highlight that our method incorporates personalized text-to-image models developed by other artists.
These models may contain inappropriate content and can be used with our method.

To address these concerns, we uphold the highest ethical standards in our research, including adhering to legal frameworks, respecting privacy rights, and encouraging the generation of positive content.
Furthermore, we believe that introducing an additional content safety checker, similar to that in Stable Diffusion~\citep{rombach2022high}, could potentially resolve this issue.

\section{Reproducibility Statement}
We provide comprehensive implementation details for the training and inference of our method in supplementary materials, aiming to enhance the reproducibility of our approach.
We also make both the code and pre-trained weights open-sourced to facilitate further investigation and exploration.

\section*{Acknowledgement}
This project is funded in part by Shanghai AI Laboratory (P23KS00020, 2022ZD0160201), CUHK Interdisciplinary AI Research Institute, and the Centre for Perceptual and Interactive Intelligence (CPIl) Ltd under the Innovation and Technology Commission (ITC)’s InnoHK.

\bibliography{iclr2024_conference}
\bibliographystyle{iclr2024_conference}

\end{document}